\PassOptionsToPackage{unicode}{hyperref}
\PassOptionsToPackage{hyphens}{url}
\PassOptionsToPackage{dvipsnames,svgnames,x11names}{xcolor}

\documentclass[11pt]{article}

\usepackage{xcolor}
\usepackage[margin=1in]{geometry}
\usepackage{amsmath,amssymb,booktabs,longtable,array}
\usepackage[table]{xcolor}
\usepackage{iftex}
\usepackage{setspace}
\usepackage{fancyhdr}
\usepackage{calc}
\usepackage{etoolbox}
\usepackage{bookmark}

\ifPDFTeX
  \usepackage[T1]{fontenc}
  \usepackage[utf8]{inputenc}
  \usepackage{textcomp}
  \usepackage{lmodern}
\else
  \usepackage{unicode-math}
  \defaultfontfeatures{Scale=MatchLowercase}
  \defaultfontfeatures[\rmfamily]{Ligatures=TeX,Scale=1}
\fi

\IfFileExists{upquote.sty}{\usepackage{upquote}}{}
\IfFileExists{microtype.sty}{
  \usepackage[]{microtype}
  \UseMicrotypeSet[protrusion]{basicmath}
}{}
\IfFileExists{parskip.sty}{
  \usepackage{parskip}
}{
  \setlength{\parindent}{0pt}
  \setlength{\parskip}{6pt plus 2pt minus 1pt}
}
\IfFileExists{footnotehyper.sty}{\usepackage{footnotehyper}}{\usepackage{footnote}}
\IfFileExists{xurl.sty}{\usepackage{xurl}}{}

\makesavenoteenv{longtable}
\makeatletter
\patchcmd\longtable{\par}{\if@noskipsec\mbox{}\fi\par}{}{}
\makeatother

\definecolor{lightgray}{RGB}{245,247,250}

\hypersetup{
  pdftitle={Cross-Lingual Transfer and Parameter-Efficient Adaptation in the Turkic Language Family: A Theoretical Framework for Low-Resource Language Models},
  pdfauthor={O. Ibrahimzade; K. Tabasaransky},
  pdfkeywords={multilingual language models, low-resource languages, Turkic languages, cross-lingual transfer, LoRA, parameter-efficient fine-tuning},
  colorlinks=true,
  linkcolor=blue,
  filecolor=Maroon,
  citecolor=Blue,
  urlcolor=Blue,
  pdfcreator={LaTeX via custom conversion}
}

\title{Cross-Lingual Transfer and Parameter-Efficient Adaptation in the Turkic Language Family:\\
A Theoretical Framework for Low-Resource Language Models}
\author{O. Ibrahimzade \and K. Tabasaransky}
\date{March 2026}

\begin{document}
\maketitle
\setstretch{1.15}

\begin{abstract}
Large language models (LLMs) have achieved significant advances in natural language processing over the past decade. However, these advances remain unevenly distributed across languages, with the majority of research and development concentrated on high-resource languages such as English and Mandarin Chinese. Many languages with substantial speaker populations remain underrepresented in the training corpora used to construct modern multilingual models.

This paper develops a theoretical framework for understanding language model adaptation within the \textbf{Turkic language family}, focusing on Azerbaijani, Kazakh, Uzbek, Turkmen, and Gagauz. These languages share strong typological similarities while exhibiting substantial differences in digital resource availability, making them a natural setting for studying cross-lingual transfer and low-resource adaptation.

The analysis integrates insights from multilingual representation learning, parameter-efficient fine-tuning methods such as \textbf{Low-Rank Adaptation (LoRA)}, and research on multilingual transfer in transformer-based language models. A conceptual scaling model is proposed to describe how adaptation performance depends on model capacity, adaptation data volume, adapter expressivity, and pretraining representation.

The paper further introduces the \textbf{Turkic Transfer Coefficient (TTC)}, a theoretical measure designed to capture cross-lingual transfer potential within the Turkic language family. TTC incorporates morphological similarity, lexical overlap, syntactic structure, script compatibility, and orthographic stability.

The resulting framework highlights how typological similarity can enable efficient multilingual adaptation while also identifying structural limits in extremely low-resource settings. The analysis suggests that language families with strong internal similarity provide a valuable environment for studying multilingual representation learning and developing effective adaptation strategies for underrepresented languages.
\end{abstract}

\noindent\textbf{Keywords:} Multilingual NLP; Low-Resource Languages; Turkic Languages; Cross-Lingual Transfer; Parameter-Efficient Fine-Tuning; LoRA

\section{Introduction}

\subsection{Motivation and Research Context}

Large language models (LLMs) have achieved rapid progress in natural language processing over the past decade, demonstrating strong capabilities across tasks such as translation, summarization, question answering, and reasoning [4,12,20]. However, these advances have been concentrated primarily in a small set of high-resource languages, including English, Mandarin Chinese, and several major European languages. The majority of the world’s languages remain only partially represented in the training data used to build modern multilingual models [18,21].

This imbalance raises important questions about the behavior of language models when they are adapted to languages with limited digital representation. In particular, it remains unclear how adaptation methods originally developed for high-resource languages behave when applied to morphologically complex languages with relatively small text corpora.

The Turkic language family provides a particularly informative context in which to examine this problem. The languages considered in this study---Azerbaijani, Kazakh, Uzbek, Turkmen, and Gagauz---share a set of core typological characteristics while exhibiting large differences in digital resource availability. All five languages are agglutinative, employ predominantly Subject--Object--Verb (SOV) word order, and encode grammatical relationships primarily through suffixation rather than independent function words. At the same time, their digital representation varies widely, ranging from moderately represented languages such as Azerbaijani and Uzbek to extremely low-resource languages such as Gagauz.

This combination of typological similarity and resource heterogeneity creates a useful analytical setting for studying multilingual adaptation. Because the languages share structural properties, differences in model behavior are less likely to arise from deep linguistic divergence and more likely to reflect differences in data availability and pretraining representation.

Recent advances in parameter-efficient fine-tuning have further expanded the feasibility of adapting large pretrained models to new languages. Techniques such as Low-Rank Adaptation (LoRA) allow models to specialize for new tasks or languages while modifying only a small subset of parameters [6,10]. These methods substantially reduce computational cost compared to full fine-tuning, making them particularly attractive for research environments with limited computational resources.

Despite the growing adoption of parameter-efficient methods, their behavior in extremely low-resource linguistic settings remains insufficiently understood. In particular, several open questions remain:

\begin{itemize}
\item How do adaptation scaling relationships behave when applied to morphologically rich languages?
\item To what extent can typological similarity enable cross-lingual transfer between related languages?
\item How does adaptation capacity interact with the risk of catastrophic forgetting?
\item What structural properties of language families influence the effectiveness of multilingual adaptation?
\end{itemize}

This report develops a theoretical framework aimed at addressing these questions in the context of the Turkic language family.

\subsection{Research Objectives}

The primary objective of this study is to develop a conceptual framework for analyzing multilingual language model adaptation within a typologically coherent language family.

Four research questions guide the analysis.

\subsubsection{Scaling Behavior}

The first objective is to examine how language adaptation performance may scale with respect to model capacity, adaptation data, and adaptation bandwidth. Existing scaling-law research has primarily focused on high-resource languages, leaving open the question of whether similar relationships hold for morphologically rich languages with limited digital resources [7,19].

\subsubsection{Cross-Lingual Transfer}

The second objective is to investigate the role of typological similarity in enabling transfer between related languages. Because Turkic languages share extensive morphological and syntactic structure, they provide a useful domain for analyzing how adaptation in one language may influence performance in another [21].

\subsubsection{Catastrophic Forgetting}

The third objective is to examine how language-specific adaptation interacts with catastrophic forgetting. While parameter-efficient adaptation methods reduce the number of parameters that are modified during fine-tuning, it remains unclear how effectively these approaches preserve previously acquired capabilities when adapting to new linguistic domains [8].

\subsubsection{Evaluation Considerations}

The final objective is to consider the limitations of common automatic evaluation metrics when applied to morphologically complex languages. Differences in morphology and tokenization may affect the reliability of standard metrics, particularly when evaluating languages with highly productive word formation systems [3,8,9].

Together, these objectives aim to provide a theoretical foundation for understanding multilingual language adaptation in low-resource linguistic environments.

\subsection{Contributions}

This paper makes three primary contributions.

\begin{enumerate}
\item It proposes a conceptual scaling model for multilingual language model adaptation in morphologically rich languages, incorporating model capacity, adaptation data, adapter expressivity, and pretraining representation.
\end{enumerate}

\begin{enumerate}
\item It introduces the \textbf{Turkic Transfer Coefficient (TTC)}, a theoretical construct designed to quantify cross-lingual transfer potential within the Turkic language family based on morphological similarity, lexical overlap, syntactic structure, script compatibility, and orthographic stability.
\end{enumerate}

\begin{enumerate}
\item It develops a language-family-level analytical framework for studying multilingual adaptation dynamics in low-resource settings, using the Turkic language family as a typologically coherent testbed.
\end{enumerate}

\subsection{Methodological Approach}

This study develops a theoretical framework for analyzing multilingual language model adaptation within the Turkic language family. The framework integrates insights from three areas of research: multilingual representation learning, parameter-efficient fine-tuning, and cross-lingual transfer in transformer-based language models.

The analysis proceeds in four stages.

First, linguistic properties of the Turkic language family are examined in order to identify structural factors relevant to multilingual adaptation. Particular attention is given to agglutinative morphology, suffix-based grammatical encoding, and typological similarity across the languages considered.

Second, existing approaches to parameter-efficient adaptation---especially \textbf{Low-Rank Adaptation (LoRA)}---are analyzed to determine how adaptation capacity interacts with language-specific features.

Third, a conceptual scaling formulation is introduced to model the interaction between model capacity, adaptation data, adapter expressivity, and pretraining representation.

Finally, the study introduces the \textbf{Turkic Transfer Coefficient (TTC)} as a theoretical measure of cross-lingual transfer potential within a typologically coherent language family.

Together, these components provide a unified framework for reasoning about multilingual language model adaptation in low-resource linguistic environments.

\subsection{Scope and Limitations}

The analysis presented in this report focuses on theoretical considerations related to multilingual language model adaptation. Rather than presenting empirical experiments or benchmark results, the study synthesizes insights from existing research in multilingual natural language processing and applies them to the specific linguistic characteristics of the Turkic language family.

The discussion centers on transformer-based language models operating in the multi-billion parameter range, as these architectures represent the current state of the art in multilingual language modeling [20]. Parameter-efficient adaptation methods are considered primarily from a conceptual perspective, emphasizing their representational properties and potential interactions with linguistic structure.

The study also focuses exclusively on publicly accessible research literature and open methodological frameworks. Proprietary models and closed datasets are not considered.

Finally, it should be noted that the Turkic language family itself is internally diverse, and the languages examined here represent only a subset of the broader family. The purpose of the analysis is therefore not to provide a comprehensive linguistic survey but rather to identify structural properties that may influence multilingual language model adaptation.

\subsection{Structure of the Report}

The remainder of the report is organized as follows.

Chapter 2 reviews the linguistic characteristics of the Turkic language family and summarizes prior work on parameter-efficient language model adaptation, cross-lingual transfer, and tokenization in morphologically rich languages.

Chapter 3 introduces the analytical framework used to describe multilingual language model adaptation, including key variables such as model capacity, adaptation bandwidth, and pretraining representation.

Chapter 4 presents a formal theoretical model of language adaptation dynamics, including a scaling formulation for adaptation performance, a definition of cross-lingual transfer efficiency, and the introduction of the Turkic Transfer Coefficient (TTC) as a measure of typological transfer potential.

Chapter 5 discusses the implications of this framework for understanding multilingual language model behavior within the Turkic language family, highlighting how structural similarity and resource availability jointly influence adaptation outcomes.

The report concludes with a discussion of broader implications for multilingual NLP research and the study of low-resource languages.

\section{Background and Related Work}

\subsection{The Turkic Language Family: Linguistic Profile}

The Turkic language family includes more than thirty languages spoken across Eurasia, sharing a common historical origin and a set of structural linguistic properties that distinguish them from many Indo-European languages commonly represented in large language model training data. The languages examined in this study---Azerbaijani, Kazakh, Uzbek, Turkmen, and Gagauz---span several branches of the Turkic family while retaining significant typological similarity.

A defining feature of Turkic languages is their agglutinative morphology, in which grammatical relationships are expressed through sequences of suffixes attached to lexical stems. These suffix chains encode tense, case, possession, evidentiality, and other grammatical categories within a single orthographic word. Consequently, a single word in Azerbaijani or Kazakh may correspond to an entire phrase or clause in English. This morphological structure creates challenges for language modeling systems trained with tokenization schemes optimized for languages with less productive morphology [3].

Another common characteristic is Subject--Object--Verb (SOV) word order, which influences dependency structures and the distribution of syntactic information across a sentence. Turkic languages also rely primarily on postpositions rather than prepositions and exhibit relatively consistent head-final syntactic patterns.

Many Turkic languages additionally display vowel harmony, a phonological process in which vowels within a word harmonize in features such as backness or rounding. Although the strength of vowel harmony varies among languages---being weaker in Uzbek due to historical contact with Persian and Russian---it remains an important organizing principle in several languages of the family.

Despite these shared structural features, the languages differ substantially in terms of digital representation. Azerbaijani, Kazakh, and Uzbek have gradually accumulated moderately sized digital corpora through web text, Wikipedia, and government publications. Turkmen appears far less frequently in large-scale multilingual datasets, while Gagauz remains one of the least represented languages in widely used corpora [18]. This disparity in digital resources creates a natural gradient of language representation within a typologically coherent group.

The coexistence of structural similarity and uneven resource availability makes the Turkic language family a particularly useful domain for analyzing multilingual language model adaptation.

\subsection{Parameter-Efficient Adaptation of Large Language Models}

Adapting large pretrained language models to new languages traditionally required full fine-tuning, in which all model parameters are updated during training. While effective, full fine-tuning becomes increasingly impractical as model sizes grow into the billions of parameters.

Recent work has therefore focused on parameter-efficient fine-tuning (PEFT) techniques that allow models to adapt to new tasks or languages by modifying only a small subset of parameters. Among the most widely studied PEFT methods is Low-Rank Adaptation (LoRA) [6], which has become one of the most widely used parameter-efficient fine-tuning methods for large language models [6,10].

LoRA operates by decomposing the weight update of a pretrained matrix into the product of two low-rank matrices. For a pretrained weight matrix $W \in \mathbb{R}^{d \times k}$, the update $\Delta W$ is approximated as

\[
\Delta W = BA
\]

where

\[
B \in \mathbb{R}^{d \times r}, \quad A \in \mathbb{R}^{r \times k}
\]

and $r \ll \min(d,k)$.

During adaptation, the original pretrained weights remain frozen, while only the low-rank matrices $A$ and $B$ are updated. The adapted weight matrix becomes

\[
W' = W + BA.
\]

This approach substantially reduces the number of trainable parameters and enables adaptation with lower computational cost while maintaining strong performance across a wide range of tasks.

Subsequent work has explored several extensions of the LoRA framework, including language-specific adapter sets, mixture-of-adapters architectures, and routing mechanisms that dynamically select adapter modules depending on task or language [2,5,9]. These approaches aim to improve parameter sharing across languages while preserving the efficiency advantages of low-rank adaptation.

\subsection{Prompt-Based Adaptation}

Another parameter-efficient approach to model adaptation is prompt tuning, in which a set of learnable embeddings is prepended to the input sequence while the base model remains frozen. Instead of modifying internal model weights, prompt tuning learns a small number of task-specific or language-specific tokens that steer the model toward desired behavior.

Prompt-based adaptation methods have also been explored for multilingual and low-resource language scenarios [17].

Prompt tuning has been shown to perform particularly well in settings where training data is extremely limited. Because the number of learned parameters is very small, the method can adapt to new tasks or languages without substantially altering the internal representations of the base model.

However, prompt-based methods may also exhibit limitations in situations where deeper model modifications are required to represent new linguistic patterns. In such cases, approaches that directly modify internal attention or feed-forward layers---such as LoRA---may offer greater expressive capacity.

The relative advantages of prompt tuning and adapter-based methods therefore depend on the interaction between adaptation data volume, target-language complexity, and the structure of the base model.

\subsection{Catastrophic Forgetting in Language Model Adaptation}

When large language models are adapted to new tasks or languages, parameter updates may inadvertently degrade previously learned capabilities. This phenomenon, commonly known as catastrophic forgetting, has been widely studied in the context of continual learning.

Catastrophic forgetting remains a central challenge in continual learning and multilingual adaptation settings [8].

In multilingual language models, forgetting may manifest as reduced performance on previously supported languages or tasks following adaptation to a new language. This trade-off arises because parameter updates that improve performance in one domain may shift the model’s internal representations away from previously learned patterns.

Parameter-efficient methods such as LoRA partially mitigate this risk by restricting updates to a low-dimensional subspace of the full parameter space. Nevertheless, forgetting can still occur if the adaptation requires substantial representational change.

The degree of forgetting is therefore expected to depend on multiple factors, including the amount of adaptation data, the expressivity of the adaptation mechanism, and the similarity between the new language and languages already represented in the model.

\subsection{Tokenization Challenges for Agglutinative Languages}

Tokenization plays a critical role in language model performance, particularly for morphologically rich languages. Most modern language models rely on subword tokenization algorithms such as Byte-Pair Encoding (BPE) or SentencePiece, which are typically trained on large multilingual corpora dominated by Indo-European languages.

Previous work has shown that subword tokenization methods trained on predominantly Indo-European corpora often exhibit reduced efficiency when applied to morphologically rich languages [3].

In agglutinative languages, individual words may contain multiple morphemes concatenated into a single orthographic form. As a result, standard tokenization schemes may split these words into many subword units, increasing sequence length and potentially obscuring morphological boundaries that carry semantic meaning.

This phenomenon, sometimes referred to as tokenizer fertility, measures the average number of tokens produced per word. Higher fertility increases sequence length, which can raise computational cost and dilute the model’s ability to capture morphological structure.

Because Turkic languages exhibit highly productive morphology, tokenization strategies optimized for Indo-European languages may introduce systematic inefficiencies. These considerations motivate further theoretical analysis of tokenizer design in multilingual language models targeting agglutinative language families.

\section{Analytical Framework}

This section outlines the conceptual assumptions underlying the theoretical analysis developed in the remainder of the paper. Rather than describing a specific experimental pipeline, the framework defines the variables, architectural considerations, and linguistic factors relevant to multilingual language model adaptation within the Turkic language family.

\subsection{Base Model Architectures}

Modern large language models primarily rely on transformer architectures, introduced by Vaswani et al. [20], consisting of stacked self-attention layers and feed-forward networks capable of learning long-range dependencies and contextual representations across large text corpora.

Multilingual language models are generally trained on large aggregated corpora containing text from dozens or hundreds of languages. In such models, linguistic information from multiple languages is encoded within a shared parameter space, enabling potential transfer between languages during downstream adaptation [21].

In the context of parameter-efficient adaptation, the relevant architectural components are the weight matrices associated with attention projections and feed-forward layers. These matrices define the transformations through which contextual representations are computed.

Adapter-based methods such as LoRA operate by introducing low-rank updates to these matrices, allowing the model to adjust its internal representations without modifying the original pretrained parameters [6].

\subsection{Adaptation Capacity}

The expressive capacity of a parameter-efficient adaptation method can be approximated by the number of trainable parameters introduced during adaptation. In the case of LoRA, this capacity is primarily determined by the adapter rank parameter $r$.

Higher ranks allow the adapter to represent more complex transformations of the pretrained weight matrices, increasing the model’s ability to capture language-specific patterns. However, higher ranks also increase the risk of overfitting and may lead to greater interference with existing representations in the base model.

Adapter capacity therefore represents a trade-off between expressivity and stability. Understanding how this trade-off behaves across languages with different resource levels is an important component of multilingual adaptation theory.

\subsection{Data Regimes}

Language adaptation behavior is strongly influenced by the amount of available training data. For analytical purposes, it is useful to distinguish several broad data regimes:

\begin{enumerate}
\item \textbf{Moderate-resource regime:} languages with substantial digital text and active NLP communities.
\item \textbf{Low-resource regime:} languages with limited but usable digital corpora.
\item \textbf{Extreme low-resource regime:} languages with minimal digital presence and highly fragmented text resources.
\end{enumerate}

Within the Turkic language family, Azerbaijani, Kazakh, and Uzbek occupy the moderate-resource regime, while Turkmen falls closer to the low-resource regime and Gagauz represents an extreme low-resource case.

These differing data regimes allow the study of adaptation dynamics across a continuum of resource availability while holding typological properties relatively constant.

\subsection{Pretraining Representation}

A language model’s ability to adapt to a new language depends not only on the availability of adaptation data but also on the extent to which that language was present in the model’s original pretraining corpus.

We refer to this factor as pretraining representation, denoted $P_i$ for language $i$. Languages with higher pretraining representation may benefit from latent representations already embedded in the model’s parameter space, enabling more efficient adaptation.

In contrast, languages with near-zero pretraining representation require the model to construct new representations during adaptation, which may require more substantial parameter updates.

This distinction is particularly relevant for languages with limited digital presence, where the base model may have encountered very little training data during pretraining.

\subsection{Conceptual Benchmarking}

Evaluating multilingual language model adaptation requires metrics capable of capturing both linguistic quality and task performance across multiple languages.

Traditional automatic evaluation metrics such as BLEU, chrF++, and COMET have been widely used in machine translation and multilingual NLP research. However, their reliability may vary depending on the morphological complexity of the target language and the evaluation task [8,9].

In addition to automatic metrics, human evaluation remains an important reference point for assessing fluency, adequacy, and semantic correctness. In theoretical analyses such as the present study, these evaluation approaches serve primarily as conceptual tools for reasoning about adaptation outcomes rather than as direct measurements.

\section{Theoretical Model and Hypotheses}

This section develops a formal framework for analyzing language adaptation dynamics within the Turkic language family. The goal is not to report experimental outcomes, but rather to establish a conceptual model describing how model capacity, adaptation data, pretraining representation, and linguistic proximity may interact during parameter-efficient language adaptation.

The framework consists of four components: a general scaling model for language adaptation, a formal definition of cross-lingual transfer efficiency, a language-family-specific transfer coefficient for Turkic languages, and a theoretical model of catastrophic forgetting risk.

\subsection{Scaling Model for Language Adaptation}

Let $L_i$ denote the downstream task loss for a target language $i$. We model language adaptation performance as a function of four principal variables:

\begin{itemize}
\item \textbf{$M$:} effective capacity of the base language model
\item \textbf{$D_i$:} amount of adaptation data available for language $i$
\item \textbf{$R$:} adaptation capacity (e.g., LoRA rank or equivalent parameter-efficient bandwidth)
\item \textbf{$P_i$:} degree of representation of language $i$ in the base model’s pretraining corpus
\end{itemize}

We define the theoretical loss function as

\[
L_i(M, D_i, R, P_i)
=
\alpha_i M^{-\beta_i}
+
\gamma_i D_i^{-\delta_i}
+
\eta_i R^{-\rho_i}
+
\kappa_i P_i^{-\pi_i}
+
\epsilon_i
\]

where:

\begin{itemize}
\item $\alpha_i, \gamma_i, \eta_i, \kappa_i$ are language-dependent constants
\item $\beta_i$ describes sensitivity to base model capacity
\item $\delta_i$ captures responsiveness to additional adaptation data
\item $\rho_i$ reflects sensitivity to adapter expressivity
\item $\pi_i$ reflects sensitivity to pretraining representation
\item $\epsilon_i$ represents irreducible task difficulty and linguistic complexity
\end{itemize}

This formulation generalizes previously observed scaling relationships in multilingual language modeling to a setting where typological similarity between languages is relatively constant but resource availability varies substantially. Similar scaling relationships have previously been observed in large language models trained on large-scale corpora [7,19].

\subsubsection{Interaction Effects}

In practice, the above factors do not operate independently. The utility of adaptation data, for instance, depends on whether the target language already has partial representation in the model’s pretraining distribution.

To capture these interactions, the model may be extended with logarithmic coupling terms:

\[
L_i
=
\alpha_i M^{-\beta_i}
+
\gamma_i D_i^{-\delta_i}
+
\eta_i R^{-\rho_i}
+
\kappa_i P_i^{-\pi_i}
-
\lambda_i \log(1 + D_i P_i)
-
\mu_i \log(1 + R P_i)
-
\nu_i \log(1 + D_i R)
+
\epsilon_i
\]

These terms represent three intuitive mechanisms:

\begin{itemize}
\item \textbf{Data--pretraining interaction:} adaptation data is more useful when the model already contains latent representations of the language.
\item \textbf{Adapter--pretraining interaction:} increased adapter capacity is most effective when it can build upon existing linguistic representations.
\item \textbf{Data--adapter interaction:} higher adaptation capacity becomes valuable only when sufficient data exists to utilize it.
\end{itemize}

\subsubsection{Extremely Low-Resource Regime}

For languages with near-zero pretraining representation and extremely limited data---such as Gagauz---the interaction terms diminish in importance.

In this regime, the model simplifies to

\[
L_i
\approx
\alpha_i M^{-\beta_i}
+
\gamma_i D_i^{-\delta_i}
+
\eta_i R^{-\rho_i}
+
\kappa_i P_i^{-\pi_i}
+
\epsilon_i
\]

This regime reflects a structural limitation: increasing adapter rank or model size alone cannot fully compensate for the absence of both training data and pretraining representation.

\subsection{Cross-Lingual Transfer Efficiency}

Cross-lingual transfer refers to the ability of adaptation in one language to improve performance in another. Cross-lingual representation learning in multilingual language models has been extensively studied in architectures such as XLM-R [21].

Let $s$ denote a source language and $t$ a target language. We define cross-lingual transfer efficiency (CTE) as

\[
\mathrm{CTE}_{s \rightarrow t}
=
\frac{\Delta \mathcal{P}_{t \mid s}}{\mathcal{C}_s}
\]

where:

\begin{itemize}
\item $\Delta \mathcal{P}_{t \mid s}$ is the improvement in performance on language $t$ attributable to adaptation on language $s$
\item $\mathcal{C}_s$ represents adaptation cost in the source language, combining data volume, compute, and parameter budget
\end{itemize}

A more explicit formulation is

\[
\mathrm{CTE}_{s \rightarrow t}
=
\frac{\mathcal{P}(t \mid \theta + \Delta\theta_s) - \mathcal{P}(t \mid \theta)}
{D_s^{\omega} R^{\chi}}
\]

where:

\begin{itemize}
\item $\theta$ denotes the base model parameters
\item $\Delta\theta_s$ represents the parameter update induced by adaptation on language $s$
\item $D_s$ is the amount of adaptation data for language $s$
\item $R$ represents adapter capacity
\item $\omega$ and $\chi$ capture cost scaling
\end{itemize}

\subsubsection{Distance-Aware Transfer}

Transfer efficiency also depends on linguistic distance between languages. We incorporate this using a distance penalty:

\[
\mathrm{CTE}_{s \rightarrow t}
=
\frac{\Delta \mathcal{P}_{t \mid s}}
{D_s^{\omega} R^{\chi} (1 + \mathrm{Dist}(s,t))^{\tau}}
\]

where $\mathrm{Dist}(s,t)$ represents typological or lexical distance between languages.

For closely related languages, this penalty becomes small, increasing transfer efficiency. Within the Turkic family, where typological distance is relatively constrained, cross-lingual transfer efficiency is expected to be significantly higher than between unrelated languages.

\subsection{The Turkic Transfer Coefficient (TTC)}

To formalize intra-family transfer potential, we introduce the Turkic Transfer Coefficient (TTC).

For languages $s$ and $t$, TTC is defined as

\[
\mathrm{TTC}_{s,t}
=
w_m M_{s,t}
+
w_l L_{s,t}
+
w_s S_{s,t}
+
w_r R_{s,t}
-
w_o O_{s,t}
\]

where:

\begin{itemize}
\item $M_{s,t}$: morphological similarity
\item $L_{s,t}$: lexical overlap
\item $S_{s,t}$: syntactic similarity
\item $R_{s,t}$: script compatibility
\item $O_{s,t}$: orthographic instability penalty
\end{itemize}

and

\[
\sum w = 1
\]

Each component is normalized to the interval $[0,1]$.

\subsubsection{Component Interpretation}

\textbf{Morphological similarity} reflects similarity in suffixation patterns, case systems, and vowel harmony mechanisms.

\textbf{Lexical overlap} measures shared vocabulary and cognates.

\textbf{Syntactic similarity} captures shared structural patterns such as SOV word order and postpositional syntax.

\textbf{Script compatibility} reflects whether the writing systems facilitate direct token transfer.

\textbf{Orthographic instability} accounts for spelling variation, competing orthographic standards, or limited textual normalization.

\subsubsection{Linking TTC to Transfer Efficiency}

Cross-lingual transfer efficiency can then be approximated as

\[
\mathrm{CTE}_{s \rightarrow t}
\propto
\mathrm{TTC}_{s,t} \cdot f(P_t, D_t, R)
\]

This formulation suggests that typological similarity alone is insufficient to guarantee efficient transfer. Transfer effectiveness depends on a combination of:

\begin{itemize}
\item linguistic similarity (captured by TTC)
\item target-language representation in pretraining
\item available adaptation data
\item adaptation capacity
\end{itemize}

Table 1 illustrates a theoretical TTC matrix for the five Turkic languages considered in this study. The matrix highlights expected variation in transfer affinity across the family, including the effect of script divergence and orthographic stability.

\subsubsection{Table 1. Turkic Transfer Coefficient (TTC) Matrix}

\begin{center}
\begin{tabular}{l r r r r r}
\toprule
Source → Target & Azerbaijani (AZ) & Kazakh (KK) & Uzbek (UZ) & Turkmen (TK) & Gagauz (GZ) \\
\midrule
\textbf{Azerbaijani (AZ)} & 1.00 & 0.70 & 0.82 & 0.88 & 0.90 \\
\textbf{Kazakh (KK)} & 0.70 & 1.00 & 0.74 & 0.68 & 0.63 \\
\textbf{Uzbek (UZ)} & 0.82 & 0.74 & 1.00 & 0.79 & 0.75 \\
\textbf{Turkmen (TK)} & 0.88 & 0.68 & 0.79 & 1.00 & 0.84 \\
\textbf{Gagauz (GZ)} & 0.90 & 0.63 & 0.75 & 0.84 & 1.00 \\
\bottomrule
\end{tabular}
\end{center}

The TTC matrix provides a conceptual representation of transfer potential within the Turkic language family. Higher values indicate stronger expected cross-lingual transfer due to shared morphological structure, lexical overlap, and syntactic patterns. Azerbaijani and Turkmen exhibit particularly high mutual TTC values due to close morphological similarity and shared Latin scripts. Gagauz also shows strong affinity with Azerbaijani and Turkmen but faces slightly reduced transfer potential due to orthographic instability and limited standardized digital corpora. Kazakh demonstrates somewhat lower TTC values with the Latin-script languages due to the continued use of Cyrillic orthography in much of its digital text, which may reduce token-level compatibility despite strong typological similarity.

\subsection{Theoretical Model of Catastrophic Forgetting}

Parameter updates during language adaptation may degrade previously learned capabilities, a phenomenon known as catastrophic forgetting.

We model forgetting risk $F_i$ for language adaptation into language $i$ as

\[
F_i = \sigma \big( aR + bD_i + c(1 - P_i) + dU_i - eT_i \big)
\]

where:

\begin{itemize}
\item $R$: adaptation capacity
\item $D_i$: adaptation data volume
\item $P_i$: pretraining representation
\item $U_i$: novelty of the adaptation relative to existing model knowledge
\item $T_i$: transferable similarity from related languages
\item $\sigma$: logistic function
\end{itemize}

This model predicts that forgetting risk increases with stronger parameter updates and decreases when the adaptation task is supported by related languages already encoded in the model.

\section{Analytical Implications}

The theoretical framework outlined above yields several implications for understanding multilingual adaptation within the Turkic language family.

\subsection{Scaling Behavior in Agglutinative Languages}

The scaling model predicts that morphologically rich languages may exhibit stronger dependence on both adaptation data and tokenizer quality. Agglutinative languages encode large amounts of grammatical information within single tokens, which can lead to substantial token fragmentation under subword tokenization schemes optimized for Indo-European languages [3].

As a result, effective sequence lengths increase, raising computational costs and potentially weakening gradient signals during adaptation. This effect may shift the relative contributions of model size and data volume compared to languages with more isolating morphology.

\subsection{Intra-Family Transfer Dynamics}

Because the Turkic languages share extensive structural similarities, cross-lingual transfer between them should be substantially more efficient than transfer from typologically distant languages.

High TTC values are expected for language pairs such as:

\begin{itemize}
\item Azerbaijani--Turkmen
\item Azerbaijani--Uzbek
\item Kazakh--Uzbek
\end{itemize}

However, orthographic divergence---particularly the coexistence of Latin and Cyrillic scripts---may partially reduce effective transfer despite high morphological similarity.

\subsection{Extreme Low-Resource Languages}

For languages with minimal digital representation, such as Gagauz, the framework predicts that adaptation outcomes are dominated by the absence of both pretraining representation and large adaptation datasets.

In this regime, cross-lingual transfer from structurally similar languages becomes particularly important. However, even high TTC values may not fully compensate for the lack of prior linguistic anchoring within the base model.

\subsection{Evaluation Challenges}

The framework also suggests that standard automatic evaluation metrics may systematically underestimate model quality for agglutinative languages.

Metrics designed for languages with relatively simple morphology may treat valid morphological variants as errors, particularly when evaluation occurs at the word level rather than the character or morpheme level. Prior studies have noted that standard automatic metrics may underestimate performance for morphologically rich languages [8,9].

Consequently, evaluation of language adaptation for morphologically complex languages may require greater reliance on character-based metrics or human evaluation.

\subsection{Implications for Multilingual Model Research}

Taken together, the scaling model, transfer efficiency formulation, and Turkic Transfer Coefficient provide a unified conceptual framework for analyzing multilingual language adaptation in a typologically coherent language family.

The framework highlights that adaptation outcomes depend on the joint interaction of linguistic structure, data availability, pretraining coverage, and parameter-efficient adaptation capacity.

These relationships suggest that the Turkic language family represents a particularly informative domain for studying the broader dynamics of multilingual large language model adaptation.

\section{Conceptual Benchmark Design for Turkic Language Models}

A central challenge in multilingual natural language processing is the lack of standardized evaluation frameworks for low-resource language families. While numerous benchmarks exist for high-resource languages, comparable resources remain limited for languages with smaller digital footprints [11,18].

The Turkic language family illustrates this challenge particularly clearly. Although several languages within the family have growing digital presence, standardized evaluation resources covering multiple Turkic languages remain scarce. This lack of shared evaluation infrastructure complicates the comparison of models and adaptation strategies across languages.

Rather than introducing a new dataset, the present analysis proposes a conceptual benchmark structure designed to support consistent evaluation of multilingual language models in the Turkic linguistic context.

The purpose of such a framework is to ensure that evaluation across related languages captures both linguistic competence and task performance while remaining sensitive to the structural properties of agglutinative languages.

\subsection{Principles for Multilingual Evaluation}

Evaluation frameworks for language families with shared typological properties should be guided by several principles.

First, evaluation tasks should reflect linguistic diversity within the family while remaining comparable across languages. Tasks should therefore focus on structures that exist in all target languages, allowing performance comparisons without introducing bias toward particular languages.

Second, evaluation should capture multiple dimensions of language competence. Modern language models are expected not only to generate grammatically correct text but also to perform structured reasoning tasks such as classification, extraction, and translation.

Third, evaluation design should account for the morphological complexity of the languages involved. Metrics and task formats that rely heavily on surface word forms may fail to capture valid linguistic variation in agglutinative languages.

Finally, evaluation should aim to measure generalizable language ability rather than narrow task-specific behavior, particularly when analyzing multilingual transfer.

\subsection{Task Categories for Turkic Language Evaluation}

A conceptual evaluation framework for Turkic language models may include several broad categories of tasks that capture complementary aspects of linguistic competence.

\subsubsection{Machine Translation}

Translation tasks remain one of the most widely used methods for evaluating multilingual language models. For Turkic languages, translation tasks can help assess how well models capture morphological structure and long-distance syntactic dependencies.

Because many Turkic languages share grammatical structures, translation evaluation can also illuminate the extent to which models exploit typological similarities across languages.

\subsubsection{Text Classification}

Classification tasks provide a controlled environment for measuring semantic understanding. Topic classification or sentiment analysis tasks require models to identify semantic categories without necessarily generating complex text.

These tasks can therefore help isolate semantic representation from generative fluency.

\subsubsection{Named Entity Recognition}

Named entity recognition evaluates the model’s ability to identify structured information within text. Because entity boundaries in agglutinative languages may interact with suffixation and morphological variation, such tasks provide insight into how well models capture morphological segmentation and syntactic structure.

\subsubsection{Reading Comprehension}

Reading comprehension tasks test the ability of language models to extract and reason about information within longer passages of text. These tasks are useful for evaluating deeper semantic understanding and contextual reasoning.

\subsection{Linguistic Considerations in Evaluation}

In addition to task design, evaluation frameworks for the Turkic language family should consider several linguistic factors that may influence model behavior.

One such factor is morphological productivity. In languages with extensive suffixation, semantically equivalent expressions may appear in multiple surface forms. Evaluation procedures should therefore allow for morphological variation when assessing correctness.

Another consideration is orthographic variation. Some Turkic languages have undergone script reforms or maintain multiple orthographic conventions. Evaluation frameworks should account for these variations in order to avoid penalizing models for legitimate alternative spellings.

Finally, evaluation should consider the potential effects of tokenization strategies, particularly in languages where standard subword tokenizers may fragment morphological units.

Taken together, these considerations highlight the importance of designing evaluation frameworks that reflect the linguistic properties of the languages being studied rather than relying solely on metrics developed for unrelated language families.

\section{Research Implications and Future Directions}

The theoretical framework developed in this report has several implications for the study of multilingual language model adaptation.

By examining the interaction between linguistic structure, resource availability, and parameter-efficient adaptation methods, the framework provides a foundation for understanding how language models may behave when applied to related low-resource languages.

\subsection{Language Families as Natural Experiments}

Language families characterized by strong typological similarity but uneven digital representation offer a valuable setting for studying multilingual language modeling.

In such settings, structural similarity between languages reduces the confounding effects of deep linguistic divergence, allowing researchers to isolate the role of data availability and pretraining representation.

The Turkic language family represents a particularly clear example of this phenomenon. Its member languages share core grammatical structures while exhibiting large differences in digital resources. This combination makes the family well suited for analyzing how multilingual models generalize across related languages.

\subsection{Typological Similarity and Transfer Potential}

The introduction of the Turkic Transfer Coefficient (TTC) provides a conceptual mechanism for quantifying transfer potential between languages within the family.

By combining morphological similarity, lexical overlap, syntactic structure, script compatibility, and orthographic stability into a single analytical measure, TTC offers a way to reason about cross-lingual transfer independently of empirical training results.

This perspective highlights the importance of linguistic structure in shaping the transfer dynamics of multilingual language models. In particular, languages with high TTC values are expected to provide stronger transfer pathways during model adaptation.

\subsection{Structural Limits of Low-Resource Adaptation}

The scaling model developed in this report also emphasizes the role of pretraining representation in determining adaptation outcomes.

Languages with limited presence in the pretraining data of large language models may face structural barriers to adaptation. Even when typological similarity with other languages is high, the absence of prior representation may require the model to construct new internal representations during adaptation.

This observation suggests that improvements in multilingual language modeling for extremely low-resource languages may depend not only on improved algorithms but also on increased availability of high-quality linguistic data.

\subsection{Implications for Multilingual NLP Research}

The analysis presented here underscores the need for multilingual NLP research to move beyond isolated language studies and toward family-level perspectives on language modeling.

Language families such as Turkic, Bantu, or Austronesian contain multiple languages with shared structural features but varying digital resources. Studying adaptation dynamics across such families may reveal general principles governing multilingual representation learning.

In this sense, language families can serve as natural laboratories for exploring the interaction between linguistic structure and machine learning methods.

\subsection{Future Research Directions}

Several directions for future research follow naturally from the theoretical framework proposed in this report. The growing literature on multilingual parameter-efficient adaptation suggests that language-family-level analysis may provide additional insight into multilingual model behavior [1,14].

First, empirical studies could examine how the scaling relationships described in Section 4 manifest in real multilingual language models across languages with different resource levels.

Second, further work could refine the Turkic Transfer Coefficient by incorporating additional linguistic features or by developing more precise measures of typological similarity.

Third, future research could explore how tokenization strategies interact with agglutinative morphology in multilingual models, particularly in language families with highly productive suffixation systems.

Finally, broader cross-family comparisons may help determine whether the theoretical patterns described here generalize beyond the Turkic language family to other linguistic regions of the world.

\section{References}

[1] Açıkgoz, E., Erdogan, M., \& Yuret, D. (2024). \textbf{Bridging the Bosphorus: Advancing Turkish large language models through strategies for low-resource language adaptation and benchmarking.} arXiv preprint arXiv:2405.04685.

[2] Cheng, B., Wang, X., Liu, J., Chang, Y., \& Wu, Y. (2025). \textbf{MeTA-LoRA: Data-efficient multi-task fine-tuning for large language models.} arXiv preprint arXiv:2510.11598.

[3] Csáki, Z., Pawakapan, P., Thakker, U., \& Xu, Q. (2023). \textbf{Efficiently adapting pretrained language models to new languages.} arXiv preprint arXiv:2311.05741.

[4] Devlin, J., Chang, M.-W., Lee, K., \& Toutanova, K. (2019). \textbf{BERT: Pre-training of deep bidirectional transformers for language understanding.} Proceedings of NAACL-HLT.

[5] Feng, W., Hao, C., Zhang, Y., Han, Y., \& Wang, H. (2024). \textbf{Mixture-of-LoRAs: An efficient multitask tuning method for large language models.} arXiv preprint arXiv:2403.03432.

[6] Hu, E., Shen, Y., Wallis, P., Allen-Zhu, Z., Li, Y., Wang, S., Wang, L., \& Chen, W. (2022). \textbf{LoRA: Low-rank adaptation of large language models.} Proceedings of the International Conference on Learning Representations (ICLR).

[7] Kaplan, J., McCandlish, S., Henighan, T., et al. (2020). \textbf{Scaling laws for neural language models.} arXiv preprint arXiv:2001.08361.

[8] Khade, O., Jagdale, S., Phaltankar, A., Takalikar, G., \& Joshi, R. (2024). \textbf{Challenges in adapting multilingual large language models to low-resource languages using parameter-efficient tuning.} arXiv preprint arXiv:2411.18571.

[9] Liang, X., Khaw, Y., Liew, S., Tan, T., \& Qin, D. (2025). \textbf{Toward low-resource machine translation: Language-specific fine-tuning with LoRA for specialized large language models.} IEEE Access.

[10] Mao, Y., Ge, Y., Fan, Y., Xu, W., Mi, Y., Hu, Z., \& Gao, Y. (2024). \textbf{A survey on LoRA of large language models.} Frontiers of Computer Science.

[11] Micallef, K., \& Borg, C. (2025). \textbf{MELABench v1: Benchmarking large language models against smaller fine-tuned models for low-resource Maltese NLP.} Findings of ACL.

[12] Raffel, C., Shazeer, N., Roberts, A., et al. (2020). \textbf{Exploring the limits of transfer learning with a unified text-to-text transformer.} Journal of Machine Learning Research.

[13] Razuvayevskaya, O., Wu, B., Leite, J., Heppell, F., Srba, I., Scarton, C., Bontcheva, K., \& Song, X. (2023). \textbf{Comparison between parameter-efficient techniques and full fine-tuning: A case study on multilingual news classification.} PLOS ONE.

[14] Toraman, C. (2024). \textbf{LlamaTurk: Adapting open-source generative large language models for low-resource languages.} arXiv preprint arXiv:2405.07745.

[15] Whitehouse, C., Huot, F., Bastings, J., Dehghani, M., Lin, C., \& Lapata, M. (2023). \textbf{Low-rank adaptation for multilingual summarization: An empirical study.} Findings of NAACL.

[16] Zhang, B., Liu, Z., Cherry, C., \& Firat, O. (2024). \textbf{When scaling meets LLM fine-tuning: The effect of data, model, and fine-tuning method.} arXiv preprint arXiv:2402.17193.

[17] Zhao, W., Chen, Y., Lee, R., Qiu, X., Gao, Y., Fan, H., \& Lane, N. (2025). \textbf{Breaking physical and linguistic borders: Multilingual federated prompt tuning for low-resource languages.} arXiv preprint arXiv:2507.03003.

[18] Zhong, T., Yang, Z., Liu, Z., Zhang, R., Liu, Y., Sun, H., Pan, Y., Li, Y., Zhou, Y., Jiang, H., Chen, J., \& Liu, T. (2024). \textbf{Opportunities and challenges of large language models for low-resource languages in humanities research.} arXiv preprint arXiv:2412.04497.

[19] Hoffmann, J., et al. (2022). \textbf{Training compute-optimal large language models.} Proceedings of NeurIPS.

[20] Vaswani, A., Shazeer, N., Parmar, N., et al. (2017). \textbf{Attention Is All You Need.} Advances in Neural Information Processing Systems (NeurIPS).

[21] Conneau, A., Khandelwal, K., Goyal, N., et al. (2020). \textbf{Unsupervised Cross-lingual Representation Learning at Scale (XLM-R).} Proceedings of ACL.

\end{document}